\documentclass{article}

\usepackage{arxiv}

\usepackage[utf8]{inputenc} 
\usepackage[T1]{fontenc}    
\usepackage{hyperref}       
\usepackage{url}            
\usepackage{booktabs}       
\usepackage{amsfonts}       
\usepackage{nicefrac}       
\usepackage{microtype}      
\usepackage{lipsum}
\usepackage{graphicx}
\graphicspath{ {./images/} }
\usepackage{subcaption} 
\usepackage{amsmath,amssymb,amsfonts}
\usepackage{cleveref}

\title{On the Effectiveness of Pretraining for Graph Combinatorial Optimization}

\author{
    David Aguado,
    Daniel Fuertes,
    Carlos R. del-Blanco,
    Fernando Jaureguizar \\
    Grupo de Tratamiento de Imágenes,
    Information Processing and Telecomunications Center,\\
    ETSI Telecomunicación, Universidad Politécnica de Madrid, 28040, Madrid, Spain\\
    \texttt{david.aguado@alumnos.upm.es, \{d.fcoiras, carlosrob.delblanco, fernando.jaureguizar\}@upm.es} \\
}

\begin{document}
\maketitle
\begin{abstract}
This paper introduces a self-supervised pretraining framework for graph combinatorial optimization specifically designed to address the nature of routing problems like the Traveling Salesman Problem. By utilizing graph contrastive learning with geometric augmentations (specifically, rotations and axial reflections) the model is forced to learn invariant structural representations and global relative distance distributions. Results demonstrate that this pretraining strategy outperforms non-pretrained models across various problem scales. Notably, the hybrid strategy (combining rotation and reflection) achieved a 6.57\% improvement in tour length for TSP1000, proving that geometric pretraining is an important inductive bias for effectively scaling neural solvers to high-dimensional instances.
\end{abstract}


\section{Introduction}
Neural Combinatorial Optimization (NCO) has recently emerged as a promising research area, leveraging deep learning to tackle classical NP-hard problems like Vehicle Routing Problems (VRPs). Among these, the Traveling Salesman Problem (TSP) \cite{bellmore1968} has become an important benchmark due to its simplicity and broad applicability. Learning-based approaches \cite{kool2019}, usually trained with Deep Reinforcement Learning (DRL), have demonstrated competitive performance compared to traditional heuristics, such as 2-opt \cite{2ops}, Christofides \cite{Christofides}, or the Lin-Kernighan algorithm \cite{Kernighan}. However, their success relies heavily on the quality of the learned graph embeddings.

In other domains, self-supervised pretraining is standard for extracting robust representations. For instance, GraphMAE \cite{mae} utilizes masked autoencoders in molecular biology, while frameworks like GCA \cite{GCA} and SimGRACE \cite{SimGRACE} leverage adaptive augmentations and contrastive learning to exploit structural connectivity in social and functional networks. However, these methods are not directly applicable to VRPs. Unlike those domains, VRP graphs are ``attribute-poor," consisting almost exclusively of 2D coordinates. Furthermore, routing requires capturing the global distribution of relative distances in fully connected graphs, rather than sparse topologies.

\begin{figure}[t!]
    \centering
    \begin{subfigure}[b]{0.325\linewidth}
        \centering
        \includegraphics[width=\linewidth]{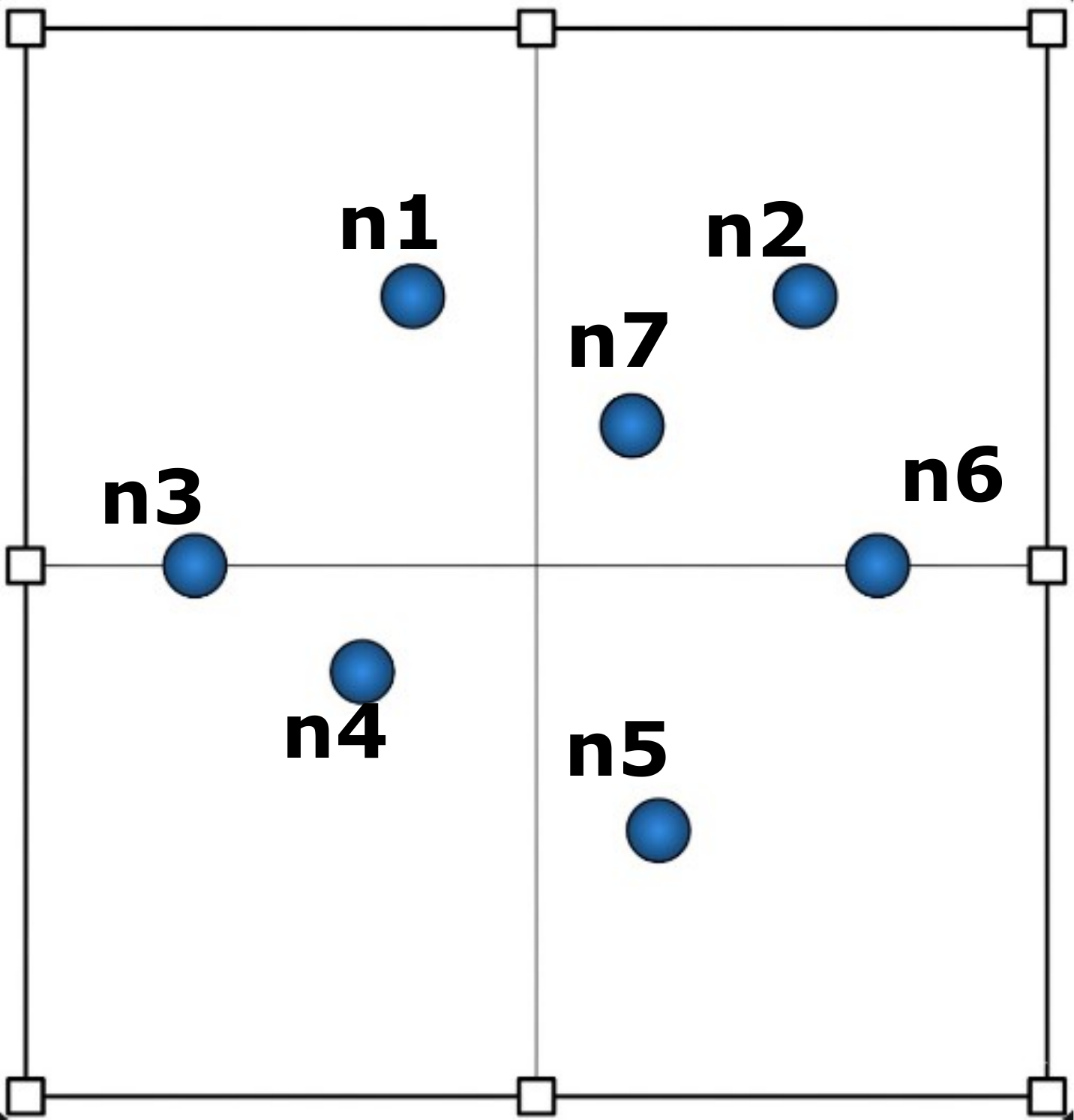}
        \caption{Original}
        \label{fig:trans_a}
    \end{subfigure}
    \hfill 
    \begin{subfigure}[b]{0.325\linewidth}
        \centering
        \includegraphics[width=\linewidth]{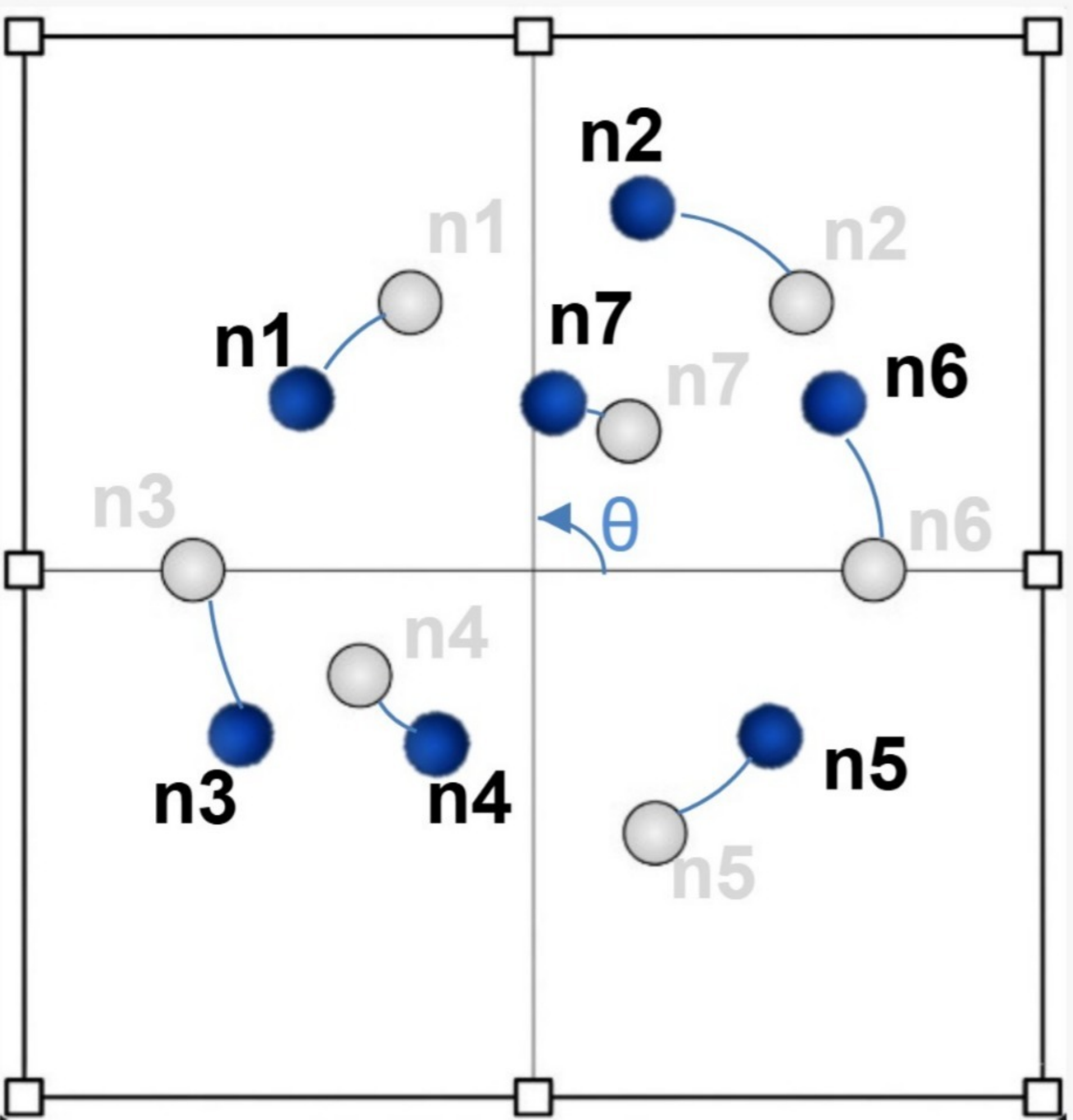}
        \caption{Rotated}
        \label{fig:trans_b}
    \end{subfigure}
    \hfill
    \begin{subfigure}[b]{0.318\linewidth}
        \centering
        \includegraphics[width=\linewidth]{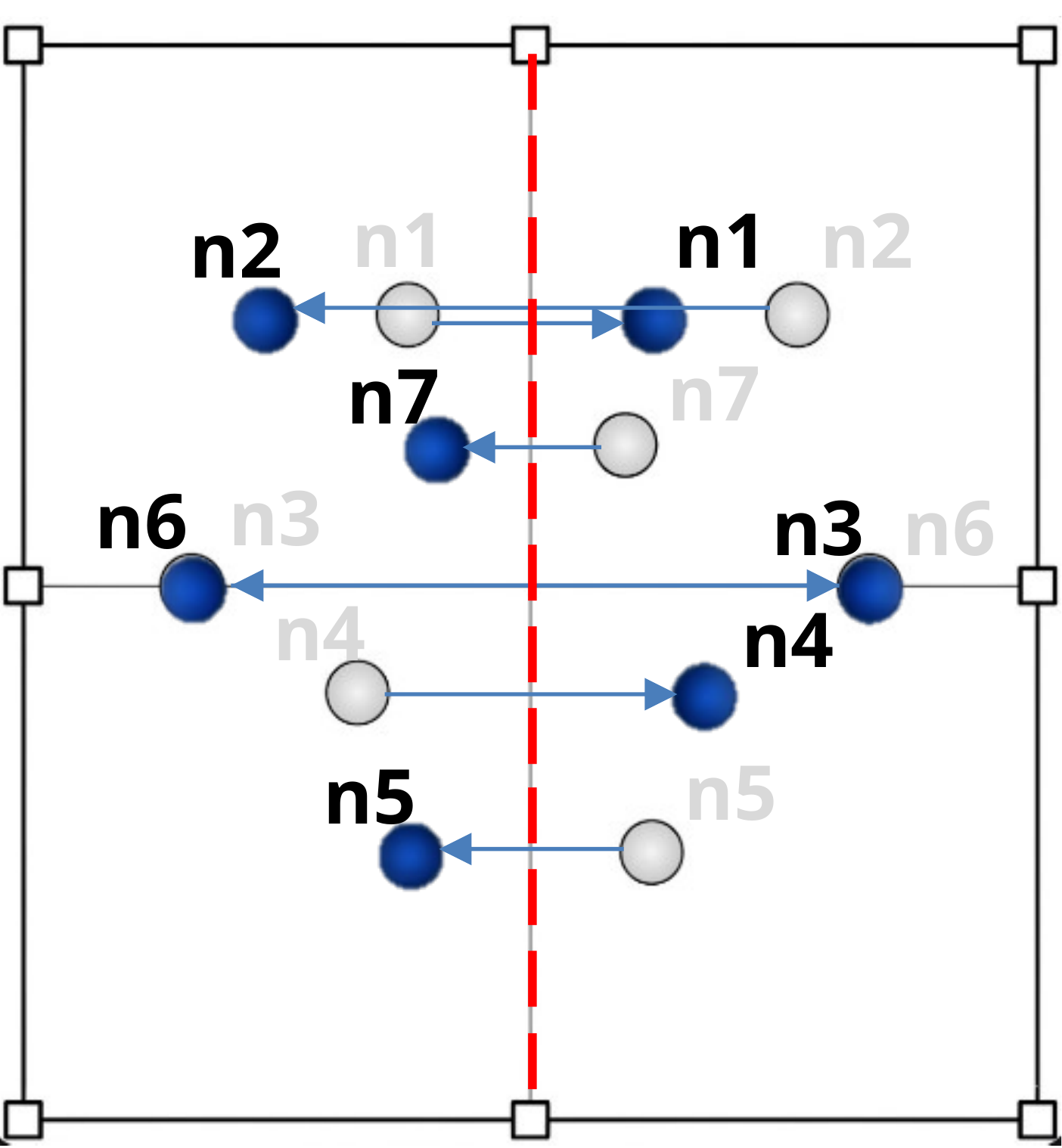}
        \caption{Reflected}
        \label{fig:trans_c}
    \end{subfigure}
    
    \caption{Geometric transformations for graph pretraining: (a) original instance, (b) rotation $f_\theta$, and (c) axial reflection $f_\phi$.}
    \label{fig:transformations_summary}
\end{figure}

To bridge this gap, we propose a pretraining framework specifically designed for the geometric nature of routing challenges. Our main contribution is a graph contrastive learning approach that employs geometric augmentations, such as rotations and axial reflections that do not change the graph topology, to force the model to learn invariant structural representations, and thus capturing the essential geometric properties of the graph representation.

\begin{table}[htbp]
    \caption{Performance comparison (average tour length $\pm$ 95\% confidence interval) using geometric pretraining. Best in bold.}
    \label{tab1}
    \centering
    \small
    \setlength{\tabcolsep}{4pt}
    \begin{center}
        \begin{tabular}{|l|c|c|c|c|c|c|}
            \hline
            
            \textbf{Pretraining} &
            \textbf{\textit{TSP20}} & \textbf{\textit{TSP50}} & \textbf{\textit{TSP100}} & \textbf{\textit{TSP200}} & \textbf{\textit{TSP500}} & \textbf{\textit{TSP1000}} \\
            \hline
            
            None &
            $3.989 \pm 0.020$ &
            $6.078 \pm 0.019$ &
            $8.796 \pm 0.031$ &
            $13.321 \pm 0.064$ &
            $24.052 \pm 0.092$ &
            $38.664 \pm 0.137$ \\
            
            Rotation &
            $3.954 \pm 0.020$ &
            $6.039 \pm 0.018$ &
            $8.666 \pm 0.033$ &
            $13.080 \pm 0.066$ &
            $23.278 \pm 0.094$ &
            $36.835 \pm 0.134$ \\
            
            Reflection &
            $\mathbf{3.953 \pm 0.020}$ &
            $\mathbf{6.030 \pm 0.019}$ &
            $8.651 \pm 0.028$ &
            $\mathbf{12.927 \pm 0.064}$ &
            $23.201 \pm 0.089$ &
            $36.694 \pm 0.132$ \\
            
            Hybrid &
            $\mathbf{3.953 \pm 0.020}$ &
            $\mathbf{6.030 \pm 0.018}$ &
            $\mathbf{8.623 \pm 0.028}$ &
            $12.936 \pm 0.066$ &
            $\mathbf{22.887 \pm 0.091}$ &
            $\mathbf{36.123 \pm 0.139}$ \\
            \hline
        
        \end{tabular}
    \end{center}
\end{table}

\section{Methodology}
The proposed framework incorporates a geometric pretraining phase based on graph contrastive learning, consisting of forcing that embeddings from equivalent node graphs are similar and vise versa. The assumption for claiming that two different node graphs are equivalent is to share the same route solution. Given a graph representing a TSP, defined by its node Euclidean distances, another equivalent graph can be obtained by isometries transformations since they preserve the distance between every pair of nodes and, therefore, the solution remains invariant. The proposed isometric transformations are rotations and axial reflections, illustrated in Fig. \ref{fig:transformations_summary}.

Specifically, for a set of nodes $\mathcal{N} = \{1, \dots, n\}$ defined in a normalized space with coordinates $\mathbf{x}_i \in [0, 1]^2 \subset \mathbb{R}^2, i \in \mathcal{N}$, we define a rotation transformation $f_\theta: \mathbb{R}^2 \to \mathbb{R}^2$ relative to the unit square's center $\mathbf{c} = [0.5, 0.5]^\top$. The rotation angle $\theta$ is sampled from the uniform distribution $\mathcal{U}(0, 2\pi)$, and The transformed coordinates are computed as follows:

\vspace{-0.2cm}
\begin{equation}
    f_\theta(\mathbf{x}_i) = \mathbf{R}_\theta(\mathbf{x}_i - \mathbf{c}) + \mathbf{c}, \quad
    \mathbf{R}_\theta = \begin{bmatrix} \cos\theta & -\sin\theta \\ \sin\theta & \cos\theta \end{bmatrix}
\end{equation}

where $\mathbf{R}_\theta \in \mathbb{R}^{2 \times 2}$ is the 2D rotation matrix.

Similarly, we define an axial reflection $f_\phi: \mathbb{R}^2 \to \mathbb{R}^2$ across an axis passing through the center $\mathbf{c}$ with an orientation angle $\phi \sim \mathcal{U}(0, \pi)$. To represent this mathematically, let $\mathbf{v} = [\cos\phi, \sin\phi]^\top$ be the unit vector defining the axis of reflection. The transformation is calculated using the Householder reflection matrix $\mathbf{S}_\phi = 2\mathbf{v}\mathbf{v}^\top - \mathbf{I}$, resulting in the following expression for the reflected coordinates:

\vspace{-0.2cm}
\begin{equation}
    f_\phi(\mathbf{x}_i) = \mathbf{S}_\phi(\mathbf{x}_i - \mathbf{c}) + \mathbf{c}.
\end{equation}

To further increase graph diversity, these functions can be applied independently or combined into a hybrid transformation $g(\mathbf{x}_i)=f_\theta(f_\phi(\mathbf{x}_i))$, which applies both operations simultaneously.

To enhance the model's adaptability across different problem scales, the pretraining dataset consists of TSP instances with variable sizes, where the number of nodes $N$ is uniformly sampled such that $N \in [20, 50]$. The encoder is pretrained over $50$ epochs with an epoch size of $128,000$ instances and a batch size of $256$ graphs. This process exposes the model to a total of $6.4 \times 10^6$ unique graph geometric configurations, providing a massive and diverse pretraining. This strategic diversity in graph size prevents the encoder from overfitting to a single graph size, fostering a more flexible structural representation that facilitates zero-shot generalization to larger, unseen instances during the evaluation phase. 

During the pretraining phase, the model encoder generates an embedding for each graph and is trained, by means of the InfoNCE contrastive loss, to maximize similarity between embeddings from equivalent graphs and minimize similarity otherwise. The InfoNCE contrastive loss is given by
\begin{equation}
\mathcal{L}_\text{InfoNCE} = -\log \frac{\exp \left( q^{\top} k{+} / \tau \right)}{\sum_{i=0}^{K} \exp \left( q^{\top} k_{i} / \tau \right)}
\end{equation}
where $q$ represents an embedding of a graph $G$, $k_{+}$ an embedding of a graph resulting from applying one of the proposed isometric transformations to $G$, and $k_{i}$ and embedding from another graph not equivalent to $G$. And $\tau$ is a temperature parameter. This loss function forces the encoder to capture relative distance distributions and connectivity patterns that are essential for the posterior training phase.

\section{Results}
\begin{figure}[htbp] 
    \centering
    \includegraphics[width=\linewidth]{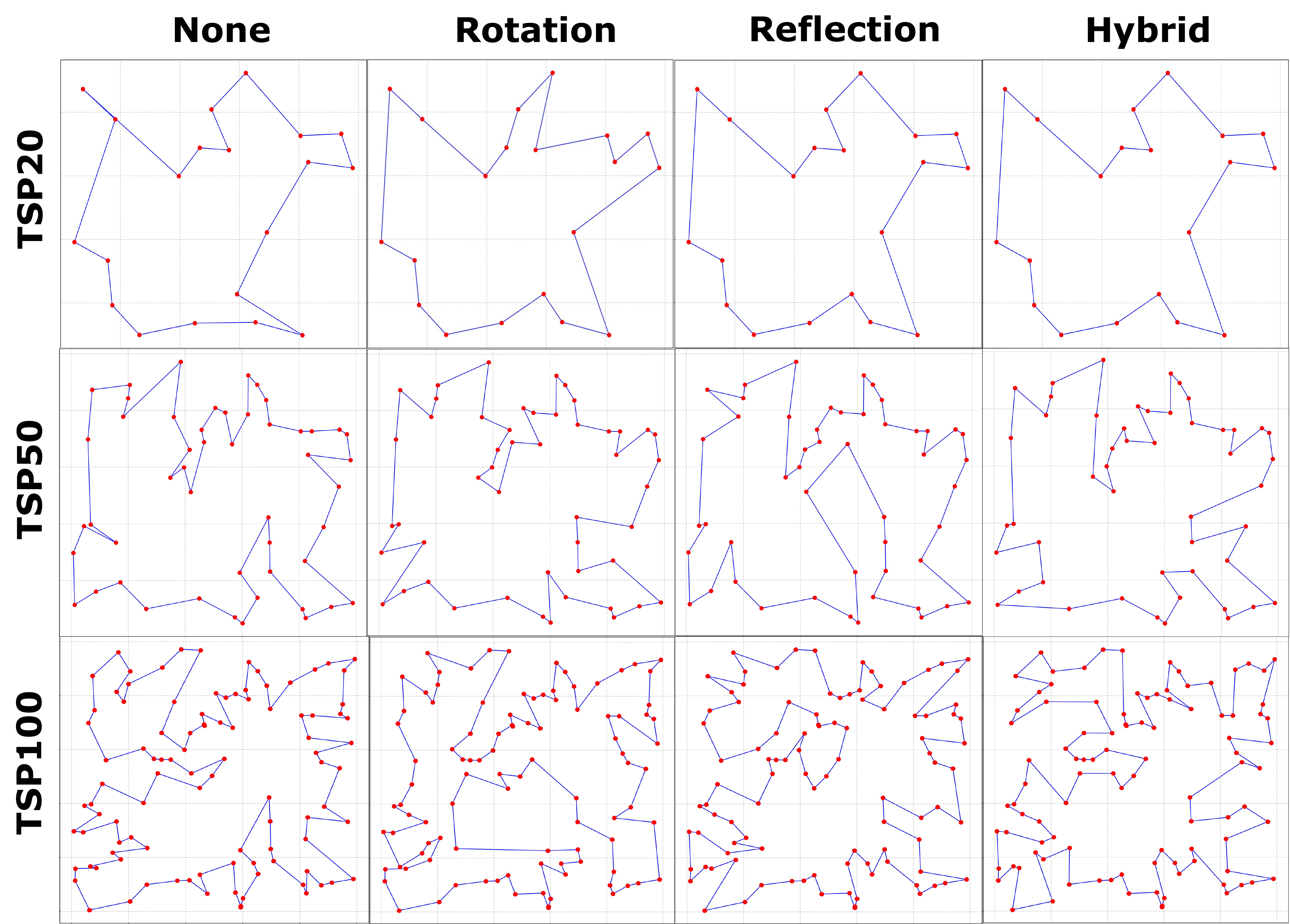}

    \caption{Examles of TSP solutions per strategy and scale.}
    \label{fig:results_matrix}
    \vspace{-0.5cm}
\end{figure}

The proposed pretraining strategy is evaluated in a Graph Convolutional Network (GCN) introduced in \cite{b4}. Qualitative results are provided in Fig. \ref{fig:results_matrix} showing a comparison of the tours generated on different scales. Observe that paths from pretrained models, especially under the hybrid strategy, exhibit fewer self-intersections.

Quantitative results in \cref{tab1} demonstrate that all pretraining variants consistently outperform the baseline model in all instance sizes. Notably, performance improves significantly as instance complexity increases, highlighting the superior scalability of our approach. For TSP20, hybrid pretarining reduces tour length by $0.90\%$, while for TSP1000, by $6.57\%$.

This growing margin indicates that, while standard DRL training is sufficient for small graphs, it struggles to generalize the underlying structural topology as the search space expands. By contrast, incorporating geometric transformations, such as rotation and axial reflection, forces the encoder to learn invariant representations, providing crucial inductive bias. This regularization allows the model to maintain a robust understanding of relative node distributions, which is critical for large-scale node instances. Finally, the hybrid approach remains the most robust, as the diversity of transformations encourages richer, more generalizable graph embeddings.

\section{Conclusions}
This paper establishes that geometric graph contrastive learning provides an effective framework for pretraining on the attribute-poor graphs inherent to the TSP. By exploiting geometric symmetries such as rotation and axial reflection, our method successfully captures invariant structural representations that significantly enhance both solution quality and model robustness compared to training from scratch. Notably, this geometric inductive bias allows the model to maintain performance as graph complexity increases, reducing the scalability gap for large-scale instances. For future research, more complex strategies specifically tailored to the nature of VRPs will be investigated.

\subsection*{Acknowledgment}
This work was supported in part by the Comunidad de Madrid under project TEC-2024/COM-322 (IDEALCVCM), in part by MCIU/AEI/10.13039/501100011033 of the Spanish Government under project PID2023148922OA-I00 (EEVOCATIONS), and in part by ``Ayudas a la Investigación para el Personal Docente e Investigador de la ETSIT-UPM (2026)'' under project ``SATURNO''. The authors would also like to thank Airbus Defence and Space for their support.

\bibliographystyle{unsrt}  
\bibliography{references}

\end{document}